\documentclass[letterpaper, 10 pt, journal, twoside]{IEEEtran}
\IEEEoverridecommandlockouts                             
\usepackage{graphics} 
\graphicspath{ {./figures/}}
\usepackage{epsfig} 
\usepackage[figurename=Fig.]{caption}

\usepackage{times} 
\usepackage{amsmath} 
\usepackage{amssymb}  
\usepackage{xcolor}
\usepackage{bm}
\usepackage{cancel}
\usepackage{color}
\usepackage{setspace}
\usepackage{wrapfig}
\usepackage{tikz}
\usetikzlibrary{shapes,arrows.meta,calc}
\usetikzlibrary{positioning}
\usepackage{caption}
\usepackage{subcaption}
\usepackage{booktabs}
\usepackage{multirow}
\usepackage{multicol}
\usepackage{soul}
\usepackage{cite}
\usepackage[normalem]{ulem}
\usepackage{xcolor}

\let\labelindent\relax 
\usepackage{enumitem}
\usepackage{tabularx}
\usepackage{courier}
\usepackage{algorithm}
\usepackage{algpseudocode} 

\usepackage{pifont}

\allowdisplaybreaks

\newcommand{\norm}[1]{\lVert #1 \rVert}
\newcommand{\dd}{\text{d}}

\newtheorem{remark}{Remark}
\newcommand{\revision}[1]{{\textcolor{black}{#1}}}
\newcommand{\revisionNew}[1]{{\textcolor{black}{#1}}}

\title{
Simultaneous Spatial and Temporal Assignment for Fast UAV Trajectory Optimization using Bilevel Optimization
}

\author{Qianzhong Chen, Sheng Cheng, and Naira Hovakimyan
\thanks{Manuscript received: December 1, 2022; Revised March 10, 2023; Accepted April 8, 2023.}
\thanks{This paper was recommended for publication by Editor Hanna Kurniawati upon evaluation of the Associate Editor and Reviewers' comments.
This work was supported by National Aeronautics and Space Administration (NASA) grant 80NSSC22M0070, National Science Foundation (NSF) under the RI grant \#2133656, and Air Force Office of Scientific Research (AFOSR) grant FA9550-21-1-0411.} 
\thanks{All the authors are with the Department of Mechanical Science and Engineering,  University of Illinois Urbana-Champaign, USA.
        {\tt\footnotesize \{qc19, chengs, nhovakim\}@illinois.edu}}%
\thanks{Digital Object Identifier (DOI): see top of this page.}
}

\begin{document}

\maketitle


\begin{abstract}
In this paper, we propose a framework for fast trajectory planning for unmanned aerial vehicles (UAVs). Our framework is reformulated from an existing bilevel optimization, in which the lower-level problem solves for the optimal trajectory with a fixed time allocation, whereas the upper-level problem updates the time allocation using analytical gradients. The lower-level problem incorporates the safety-set constraints (in the form of inequality constraints) and is cast as a convex quadratic program (QP). Our formulation modifies the lower-level QP by excluding the inequality constraints for the safety sets, which significantly reduces the computation time. The safety-set constraints are moved to the upper-level problem, where the feasible waypoints are updated together with the time allocation using analytical gradients enabled by the OptNet. We validate our approach in simulations, where our method's computation time scales linearly with respect to the number of safety sets, in contrast to the state-of-the-art that scales exponentially.
\end{abstract}

\section{Introduction}

\IEEEPARstart{T}{rajectory} planning has long been a critical problem in robotics. 
As trajectory planning affects the quality of robots’ motion to a great extent, people have long been investigating this problem for various kinds of robots~\cite{gasparetto20151}. Pioneered by Mellinger et al. \cite{Mellinger20111}, the minimum-snap method has been dominantly used in  trajectory planning \revisionNew{of UAVs with differentially flat dynamics (e.g., quadrotors)}. Such a method takes the snap of the whole trajectory as cost and generates the planning problem in the form of a quadratic program (QP). Meanwhile, various equality or inequality constraints with different requirements on the trajectory  can be added to the QP, making it convenient for addressing more complex tasks.

One of the most important features of a trajectory planning algorithm is that it needs to quickly generate a feasible trajectory for the given UAV. The computational efficiency is not a significant issue if the optimization problem is convex, e.g., the minimum-snap formulation~\cite{Mellinger20111} with fixed time allocation to the adjustable waypoints.

However, the optimization problem normally becomes non-convex when the time allocation is included as an optimization variable. Since time allocation can largely affect the quality of the planned trajectory, researchers are searching for more efficient ways to optimize time allocation for trajectory planning. Early trials have used heuristics \cite{Liu20171} and decoupling methods~\cite{Gao20182}, although they both can lead to inefficient trajectories and burdensome computation time. Mellinger et al. \cite{Mellinger20111} propose to solve this problem by gradient descent, which can be seen as the prototype of the bilevel optimization method, though the finite difference is used to approximate the gradient. Sun et al.~\cite{Sun20201} finalize the idea of using bilevel optimization to determine time allocation using analytical gradients. In their design, the lower-level problem optimizes the trajectory with fixed time allocation, which is generated by the upper-level problem that optimizes the time allocation using analytical gradients.  

Another feature that burdens computational efficiency is the requirement of collision avoidance. To fulfill the collision-free requirement, a flight corridor, which is a space with safety assurance, must be specified in advance. Ways to plan a trajectory with flight corridors vary, e.g., sampling method \cite{Mellinger20111}, two-staged planning strategy~\cite{chen20161}, Bernstein polynomials \cite{Gao20181,Sun20201}. Notwithstanding, all of the methods mentioned above are characterized as inequality constraints, introducing burdens on QP’s computation time as a QP can be solved in a much shorter time when equality constraints are involved only.

We propose a bilevel optimization framework to plan a trajectory subject to variable time allocation and safety-set constraints. Our formulation is similar to that of~\cite{Sun20201}: the lower-level problem is convex and solved for the global minimum; the upper-level problem is non-convex, and a new ``solution'' is obtained by one-step gradient descent to reduce the optimization cost.
The uniqueness of our approach is that we exclude the inequality constraint for the safety set in the lower-level problem, making it a QP with equality constraints only. Specifically, the feasible waypoints in the safety set originally characterized by the inequality constraints are fixed in the lower-level problem. Such a formulation can significantly reduce the computation time on the lower-level QP since it can be turned into an unconstrained optimization problem. 
Correspondingly, the upper-level problem will adjust the time allocation and the feasible waypoints using analytical gradients that are efficiently obtained via OptNet~\cite{Amos20171}. Compared with the upper-level problem formulation in~\cite{Sun20201}, where only the time allocation is updated via analytical gradients, the extra computation time to obtain the gradient for the spatial update (of the feasible waypoints) is negligible. This comparison concludes the major reason why our formulation can significantly reduce the computation time of that in~\cite{Sun20201}, where we validated the comparison in simulations with a scalability test. While both methods minimize the cost to a similar level, our approach's computation time scales linearly with the number of waypoints, whereas the one by~\cite{Sun20201} scales exponentially.

\textbf{Our contributions are summarized as follows}: 1. We reformulate the bilevel optimization in~\cite{Sun20201} by excluding the inequality constraints in the lower-level problem, which leads to drastically reduced computation time for an optimal solution. 2. We use analytical gradients in the upper-level problem to simultaneously update the spatial and temporal assignment, which is more accurate and efficient than using finite-difference methods or other numerical approximations.

The remainder of the paper is organized as follows: Section~\ref{sec: related work} reviews the related work. Section~\ref{sec: tech background} introduces the background of trajectory planning and the existing bilevel optimization formulation. Section~\ref{sec: method} shows our formulation of the bilevel optimization, with simulations results given  in Section~\ref{sec: simulation} showing the advantage of our method in drastically reducing the computation time. Finally, Section~\ref{sec: conclusion} summarizes the paper and discusses future work.

\section{Related Work}\label{sec: related work}

\subsection{Trajectory Planning}

A general introduction to trajectory planning is provided in \cite{gasparetto20151}. The authors of \cite{Mellinger20111} first introduce the method of minimum-snap that leverages quadrotors' differential flatness and apply monomial polynomial parametrization to the planning trajectory. The method uses snap information to build the cost function and formulate the optimization problem in the form of a convex QP. The problem can be solved efficiently with off-the-shelf solvers, making it possible for real-time onboard motion planning. Moreover, the QP formulation permits more features of trajectory planning to be integrated using equality or inequality constraints. Based on this formulation, the authors of \cite{Gao20181} replace monomial polynomials with Bernstein polynomials to confine the whole trajectory in a convex hull with safety assurance. Another variant uses B-spline instead of monomial polynomials to include curvature constraints on the planned trajectory~\cite{Kano20181}.

\subsection{Time Allocation}

As mentioned above, trajectory planning in the form of piecewise polynomials is a well-studied problem when the temporal assignment is fixed. However, as temporal assignment determines the coefficients in the cost function, it affects the resultant optimal trajectory to a great extent. People have long found it difficult to efficiently obtain an optimal time allocation, especially when planning the trajectory in a complex environment. The authors of~\cite{Richter20161} formulate the problem in the QP form with a time penalty and conduct gradient descent to iteratively find a better time allocation. However, the computation time of the method becomes the major obstacle to widely using it in real-time trajectory planning. Heuristics methods \cite{Liu20171} have also been used to achieve better time allocation. Notwithstanding, such methods do not guarantee optimal time allocation and can lead to suboptimal trajectories. Another strategy to find optimal time allocation is to decouple trajectory planning into two problems: a spatial one and a temporal one, and then make use of convex second-order conic program (SOCP) reformulation to find optimal time allocation \cite{Gao20182}. However, this method may be infeasible when the initial and final states are not static.

\subsection{Bilevel Optimization}

A bilevel optimization is a type of mathematical program where one optimization problem (the lower-level optimization problem) is embedded inside another optimization problem (the upper-level optimization problem), and the lower-level optimization problem is constrained by the upper-level optimization problem \cite{Sinha20181}. The hierarchical nature of bilevel or multi-level optimization makes it rather appropriate for solving complex optimization problems with more than one level, i.e., the coefficients of the lower level’s optimization problem are dependent on the upper level’s optimization results. In robotics, bilevel optimization has been deployed in optimal control \cite{Landry20191} and various kinds of robots’ motion planning \cite{Stouraitis20201,Menasri20151,Farshidian20171}. Sun et al. \cite{Sun20201} present a method using bilevel optimization to conduct trajectory optimization with optimal time allocation. The time allocation is solved in the upper-level problem and passed to the lower-level one as a parameter. The analytical gradients 
of the optimal cost with respect to the time allocation are obtained with Karush–Kuhn–Tucker~(KKT) conditions of the lower-level problem. Using the analytical gradient with line search, the cost in the upper-level problem is reduced, and the updated time allocation updates the parameters in the lower-level problem to generate the trajectory. In the bilevel optimization problem, the derivation of gradients is crucial. One method of obtaining gradients is by analyzing the sensitivity of the optimization problem~\cite{Pirnay20121}. A similar approach for obtaining the analytical gradient of the solution of a QP with respect to the QP's parameters via implicit differentiation over the KKT condition is shown in \cite{Amos20171}.

\section{Background} \label{sec: tech background}

We use piecewise polynomials of time to parameterize the UAV trajectories. \revisionNew{Since the} differential flatness of \revisionNew{UAVs'} dynamics \revisionNew{excludes the necessity of explicitly enforcing dynamics (e.g., quadrotors~\cite{Mellinger20111})}, we can characterize the trajectories as a smooth curve in the space of flat outputs $\boldsymbol{\sigma}: [t_0,t_m] \rightarrow \mathbb{R}^3 \times SO(2)$ as $\boldsymbol{\sigma}(t) = [x(t),y(t),z(t),\psi(t)]^\top$, which contains the coordinates of the vehicle's center of mass in the world coordinate system $\boldsymbol{r} = [x,y,z]^\top$ and yaw angle $\psi$. Consider $m$ piecewise polynomials that characterize the entire trajectory, where each piece is an $n$-th order polynomial, i.e., for $i \in \{1,2,\dots,m \}$,
\begin{equation}
    \boldsymbol{\sigma}_i(t) = \sum_{k=0}^n \boldsymbol{\sigma}_{ik} t^k, \quad t \in [t_{i-1},t_i), \ \boldsymbol{\sigma}_{ik} \in \mathbb{R}^4.
\end{equation}

The problem is to find the polynomial coefficients $\{\boldsymbol{\sigma}_{ik}\}_{i \in \mathcal{I}, k \in \mathcal{K}}$ and the temporal assignment $\boldsymbol{T} = [t_1,t_2,\dots,t_{m-1}]^\top$ such that the following cost function is minimized
\begin{equation}\label{eq: cost function}
    \sum_{i=0}^{m-1} \int_{t_i}^{t_{i+1}} \mu_r \norm{\frac{\dd^{k_r} \boldsymbol{r}(t) }{\dd t^{k_r}}}^2 + \mu_{\psi} \norm{\frac{\dd^{k_{\psi}} \psi(t) }{\dd t^{k_{\psi}}}}^2 \dd t,
\end{equation}
where $k_r$ and $k_{\psi}$ refer to the order of derivative of the coordinate $\boldsymbol{r}$ and yaw angle $\psi$, respectively. For the minimum-snap cost in~\cite{Mellinger20111}, $k_r=4$ and $k_{\psi}=2$ are used.
The constraints include convex set $\mathcal{C}_i \subseteq \mathbb{R}^3$ of adjustable waypoints at time instances, i.e.,
\begin{equation}\label{eq: feasible position constraint}
    \boldsymbol{r}_1(t_{0}) \in \mathcal{C}_0, \ \boldsymbol{r}_i(t_i) \in \mathcal{C}_i, \ i \in \{1,2,\dots,m\}.
\end{equation}
The sets $\{\mathcal{C}_i\}_{i=0:m}$ serve as one type of safety constraint, e.g., each set can be a corner connecting two corridors such that the UAV must pass through the set to ensure safety. (The usage of Bernstein polynomials can characterize the corridors as safety sets~\cite{Gao20181,Sun20201}, which is out of the scope of this paper and will not be discussed.)

Without loss of generality, the feasible sets $\mathcal{C}_i$'s are polytopes, yielding simple characterization by linear inequalities. The continuity between adjacent polynomials is enforced by
\begin{equation}
    \frac{\dd^{k_1} \boldsymbol{r}_i(t_i) }{\dd t^{k_1}} = \frac{\dd^{k_1} \boldsymbol{r}_{i+1}(t_i) }{\dd t^{k_1}}, \    \frac{\dd^{k_2} \psi_i(t_i) }{\dd t^{k_2}} = \frac{\dd^{k_2} \psi_{i+1}(t_i) }{\dd t^{k_2}}, \label{eq: continuity constraint}
\end{equation}
for $i \in \{1,2,\dots,m-1\}$, $k_1 \in \{0,1,\dots, k_r\}$, and $k_2 \in \{0,1,\dots,k_{\psi}\}$.
Finally, the temporal constraint requires the time allocation to satisfy
\begin{equation}\label{eq: time constraint}
    \boldsymbol{T} \in \mathcal{T} = \{[t_1,\dots,t_{m-1}]^\top |t_0 < t_1 <  \dots <t_{m-1} < t_m\}.
\end{equation}

The optimization problem seeks to find the polynomial coefficients $\boldsymbol{\sigma}$ and temporal assignment $\boldsymbol{T}$ that minimize the objective function in \eqref{eq: cost function} subject to the constraints in \eqref{eq: feasible position constraint}--\eqref{eq: time constraint}. Since polynomials are highly structured basis functions with coefficients captured by $\boldsymbol{\sigma}$, the problem can be conveniently presented as 
\begin{equation}\label{prob: initial formulation}
    \begin{aligned}
    & \underset{\boldsymbol{\sigma},\boldsymbol{T} \in \mathcal{T}}{\text{minimize}} && J(\boldsymbol{\sigma},\boldsymbol{T}) = \boldsymbol{\sigma}^\top P(\boldsymbol{T}) \boldsymbol{\sigma} + \boldsymbol{\sigma}^\top \boldsymbol{q}(\boldsymbol{T}) \\
    & \text{subject to} && G(\boldsymbol{T})\boldsymbol{\sigma} \preceq \boldsymbol{h},\\
    & && A(\boldsymbol{T}) \boldsymbol{\sigma} = \boldsymbol{b},
    \end{aligned}
    \tag{P}
\end{equation}
where $\boldsymbol{\sigma}$ denotes a permutation of all polynomial coefficients
$\{\boldsymbol{\sigma}_{ik}\}_{i \in \mathcal{I}, k \in \mathcal{K}}$. The inequality constraint $G(\boldsymbol{T}) \boldsymbol{\sigma} \preceq \boldsymbol{h}$ associates with the safety-set constraint~\eqref{eq: feasible position constraint}, whereas the equality constraint $A(\boldsymbol{T})=\boldsymbol{b}$ associates with the continuity constraints~\eqref{eq: continuity constraint}. All the coefficients $P(\boldsymbol{T})$, $\boldsymbol{q}(\boldsymbol{T})$, $G(\boldsymbol{T})$, $\boldsymbol{h}$, $A(\boldsymbol{T})$, and $\boldsymbol{b}$ are derived based on the objective function~\eqref{eq: cost function} and the constraints~\eqref{eq: feasible position constraint} and~\eqref{eq: continuity constraint}. The matrix $P(\boldsymbol{T})$ is positive semi-definite.

In general, the problem~\eqref{prob: initial formulation} is a non-convex problem. Therefore, bilevel optimization has been widely applied to solve the problem \eqref{prob: initial formulation}, e.g., \cite{Mellinger20111, Richter20161,Sun20201}. The bilevel optimization has an upper-level problem:
\begin{equation}\label{prob: general upper level formulation}
    \begin{aligned}
    & \underset{\boldsymbol{T} \in \mathcal{T}}{\text{minimize}} && J(\boldsymbol{\sigma}^*,\boldsymbol{T})\\
    & \text{subject to} && \boldsymbol{\sigma}^*(\boldsymbol{T}) \in \underset{\boldsymbol{\sigma}}{\text{argmin}}\{ J(\boldsymbol{\sigma},\boldsymbol{T}): \boldsymbol{\sigma} \in \mathcal{F} \},
    \end{aligned}
\end{equation}
where the $\mathcal{F}$ denotes the feasible set of $\boldsymbol{\sigma}$ such that $\mathcal{F} = \{\boldsymbol{\sigma}: G(\boldsymbol{T})\boldsymbol{\sigma} \preceq \boldsymbol{h}, A(\boldsymbol{T}) \boldsymbol{\sigma} = \boldsymbol{b}\}$. Embedded in the upper-level problem, the lower-level problem is
\begin{equation}\label{prob: general lower level formulation}
    \begin{aligned}
    & \underset{\boldsymbol{\sigma}}{\text{minimize}} && J(\boldsymbol{\sigma},\boldsymbol{T})\\
    & \text{subject to} && G(\boldsymbol{T})\boldsymbol{\sigma} \preceq \boldsymbol{h},\\
    & && A(\boldsymbol{T}) \boldsymbol{\sigma} = \boldsymbol{b}.
    \end{aligned}
\end{equation}
Note that in the lower-level problem, the time assignment $\boldsymbol{T}$ is a parameter of the problem with a known value (obtained via solving~\eqref{prob: general upper level formulation}). Hence, \eqref{prob: general lower level formulation} reduces to the minimum-snap formulation in~\cite{Mellinger20111} and turns into a quadratic program, which is convex and efficiently solvable. The challenges come from the upper-level optimization~\eqref{prob: general upper level formulation}, which is non-convex. A typical approach in solving \eqref{prob: general upper level formulation} is to use gradient descent, where the temporal assignment $\boldsymbol{T}$ is iteratively updated. In~\cite{Mellinger20111}, the descent direction is obtained by computing the directional derivative along unit vectors. In~\cite{Sun20201}, the descent direction is the gradient of the optimal lower-level cost $\nabla_{\boldsymbol{T}}J^*(\boldsymbol{\sigma}^*(\boldsymbol{T}),\boldsymbol{T})$, which is computed as an analytical gradient by implicit differentiation of the KKT condition of the low-level problem~\eqref{prob: general lower level formulation}.

\section{Method} \label{sec: method}

In this section, we describe our reformulation of the bilevel-optimization problem based on~\cite{Sun20201}. 
Our reformulation can drastically reduce the computation time to that of~\cite{Sun20201} by reducing the computation time of the lower-level QP problem. Such a QP problem is reported in~\cite{Sun20201} to dominate the computation time despite being convex. We keep equality constraints only in the lower-level problem, which turns into an unconstrained problem that can be solved in a shorter time. 

Our formulation starts by introducing a new optimization variable, which allows decomposing the inequality constraint $G(\boldsymbol{T})\boldsymbol{\sigma} \preceq \boldsymbol{h}$ into an equality constraint and an inequality constraint. Specifically, let the adjustable waypoint $\boldsymbol{\xi}_i \in \mathbb{R}^3$ be the new optimization variable such that $\boldsymbol{\xi}_i \in \mathcal{C}_i$ for $i \in \{0,1,\dots,m\}$.
Then the constraint \eqref{eq: feasible position constraint} breaks down to
\begin{equation}\label{eq: introduce adjustable waypoint}
    \boldsymbol{\xi}_0 = \boldsymbol{r}_1(t_0), \quad\boldsymbol{\xi}_0 \in \mathcal{C}_0, \quad \boldsymbol{\xi}_i = \boldsymbol{r}_i(t_i), \quad \boldsymbol{\xi}_i \in \mathcal{C}_i,
\end{equation}
for $i \in \{1,\dots,m\}$. 
Correspondingly, the inequality constraint $G(\boldsymbol{T}) \boldsymbol{\sigma} \preceq \boldsymbol{h}$ turns into 
\begin{equation}\label{eq: decomposition of the inequality constraint to split into upper/lower problems}
    C(\boldsymbol{T}) \boldsymbol{\sigma} = \boldsymbol{\xi},\quad R \boldsymbol{\xi} \preceq \boldsymbol{s},
\end{equation}
where $\boldsymbol{\xi}^\top =[\boldsymbol{\xi}_0^\top \ \boldsymbol{\xi}_1^\top \ \dots \ \boldsymbol{\xi}_m^\top ] $. The matrix $C(\boldsymbol{T})$ evaluates the positions of the trajectory on the allocated time instances in $\boldsymbol{T}$, whereas the inequality $R \boldsymbol{\xi} \preceq \boldsymbol{s}$ is the half-space representation of the polytopes $\{\mathcal{C}_i\}_{i=0:m}$.
With the newly introduced variable and constraints, problem \eqref{prob: initial formulation} turns into
\begin{equation}\label{prob: our formulation}
    \begin{aligned}
    & \underset{\boldsymbol{\sigma},\boldsymbol{\xi},\boldsymbol{T} \in \mathcal{T}}{\text{minimize}} && J(\boldsymbol{\sigma},\boldsymbol{T}) = \boldsymbol{\sigma}^\top P(\boldsymbol{T}) \boldsymbol{\sigma} + \boldsymbol{\sigma}^\top \boldsymbol{q}(\boldsymbol{T}) \\
    & \text{subject to} && R\boldsymbol{\xi} \preceq \boldsymbol{s},\\
    & && C(\boldsymbol{T}) \boldsymbol{\sigma} = \boldsymbol{\xi},\\
    & && A(\boldsymbol{T}) \boldsymbol{\sigma} = \boldsymbol{b}.
    \end{aligned}
    \tag{AP}
\end{equation}
The introduction of the new variable $\boldsymbol{\xi}$ and additional constraints preserves the equivalence between \eqref{prob: initial formulation} and \eqref{prob: our formulation}: from a solution of one problem, a solution of the other is readily found, and vice versa~\cite{boyd2004convex}. Similar to \eqref{prob: initial formulation}, problem \eqref{prob: our formulation} also can be solved using bilevel optimization. However, the formulation of \eqref{prob: our formulation} permits only keeping the equality constraints in the lower-level optimization:
\begin{equation}\label{prob: our lower level formulation}
    \begin{aligned}
    & \underset{\boldsymbol{\sigma}}{\text{minimize}} && J(\boldsymbol{\sigma},\boldsymbol{T})\\
    & \text{subject to} && C(\boldsymbol{T}) \boldsymbol{\sigma} = \boldsymbol{\xi},\\
    & && A(\boldsymbol{T}) \boldsymbol{\sigma} = \boldsymbol{b},
    \end{aligned}
    \tag{LLP}
\end{equation}
where the spatial and temporal assignments, $\boldsymbol{\xi}$ and $\boldsymbol{T}$, are generated from the upper-level optimization:
\begin{equation}\label{prob: our upper level formulation}
    \begin{aligned}
    & \underset{\boldsymbol{\xi} \in \mathcal{X},\boldsymbol{T} \in \mathcal{T}}{\text{minimize}} && J(\boldsymbol{\sigma}^*,\boldsymbol{T})\\
    & \text{subject to} && \boldsymbol{\sigma}^*(\boldsymbol{\xi},\boldsymbol{T}) \in \underset{\boldsymbol{\sigma}}{\text{argmin}}\{ J(\boldsymbol{\sigma},\boldsymbol{T}): \boldsymbol{\sigma} \in \mathcal{F}(\boldsymbol{\xi},\boldsymbol{T}) \}.
    \end{aligned}
    \tag{ULP}
\end{equation}
Here, the set $\mathcal{F}(\boldsymbol{\xi},\boldsymbol{T})$ denotes the feasible set of $\boldsymbol{\sigma}$ such that $\mathcal{F}(\boldsymbol{\xi},\boldsymbol{T}) = \{\boldsymbol{\sigma}: C(\boldsymbol{T})\boldsymbol{\sigma} = \boldsymbol{\xi}, A(\boldsymbol{T}) \boldsymbol{\sigma} = \boldsymbol{b}\}$ and $\mathcal{X}$ denotes the feasible waypoints such that $\mathcal{X} = \{\boldsymbol{\xi}: R \boldsymbol{\xi} \preceq \boldsymbol{s}\}$. 

\revision{
\begin{remark}
Our formulation can be extended to include dynamic constraints of a UAV, e.g., velocity and acceleration limits. Such constraints can be cast as convex sets $\mathcal{C}_i^v$ and $\mathcal{C}_i^a$ for feasible velocities and accelerations, respectively. In other words, we have
\begin{align}
    & \boldsymbol{v}_1(t_{0}) \in \mathcal{C}_0^v, \ \boldsymbol{v}_i(t_i) \in \mathcal{C}_i^v, \
 \boldsymbol{a}_1(t_{0}) \in \mathcal{C}_0^a, \ \boldsymbol{a}_i(t_i) \in \mathcal{C}_i^a,
\end{align}
for $i \in \{1,2,\dots,m\}$ (analogous to \eqref{eq: feasible position constraint} for the feasible waypoints), which can be augmented to the inequality constraints $G(\boldsymbol{T})\boldsymbol{\sigma} \preceq \boldsymbol{h}$ in \eqref{prob: initial formulation}. Subsequently, the decomposition in~\eqref{eq: decomposition of the inequality constraint to split into upper/lower problems} will include the augmented velocity and acceleration assignments. Therefore, our design of lower- and upper-level problems also applies to the dynamic constraints, where the upper-level one determines the temporal allocation and augmented spatial assignments (for position, velocity, and acceleration), and the lower-level one determines the polynomial coefficients to meet these assignments.
\end{remark}
}

A few notes follow our construction of the lower-upper-level problems. First, the lower-level optimization is still a convex problem: a quadratic program with equality constraints only. The insight here is to significantly reduce the computation time of the quadratic program by keeping the equality constraints only. \revision{Such a problem can be reduced to an equivalent unconstrained problem by introducing Lagrangian multipliers, which reduces to solving a system of linear equations ~\cite{boyd2004convex}.} 
Second, the upper-level problem, despite different forms than in~\cite{Mellinger20111, Richter20161,Sun20201}, is generally hard to solve due to its non-convexity. Coordinate-descent may be applied to successively solve \eqref{prob: our lower level formulation} and \eqref{prob: our upper level formulation} in a row despite the fact that the computation time can be forbidden for real-time planning. Instead, we use gradient-descent on \eqref{prob: our upper level formulation} to only update on $\boldsymbol{\xi}$ and $\boldsymbol{T}$ once \eqref{prob: our lower level formulation} is solved. Consequently, the updated values of $\boldsymbol{\xi}$ and $\boldsymbol{T}$ are plugged into~\eqref{prob: our lower level formulation} as parameters. In order to update $\boldsymbol{\xi}$ and $\boldsymbol{T}$, we need $\partial J^* / \partial \boldsymbol{\xi}$ and $\partial J^* / \partial \boldsymbol{T}$ to update $\boldsymbol{\xi}$ and $\boldsymbol{T}$ by
\begin{equation}\label{eq: projected gradient descent to update the parameters}
    \boldsymbol{\xi} \leftarrow P_{\mathcal{X}} \left( \boldsymbol{\xi} - \alpha_1 (\frac{\partial J^*}{\partial \boldsymbol{\xi}} )^\top \right), \ \boldsymbol{T} \leftarrow P_{\mathcal{T}} \left( \boldsymbol{T} - \alpha_2 (\frac{\partial J^*}{\partial \boldsymbol{T}})^\top\right),
\end{equation}
where $\alpha_1,\alpha_2 > 0$ are the user-selected step sizes and $P_{\mathcal{Y}}(y)$ denotes the projection operator~\cite{parikh2014proximal} that projects $y$ onto the set $\mathcal{Y}$. \revision{The step sizes $\alpha_1$ and $\alpha_2$ dominate the update of the spatial assignment and time allocation, respectively. Since the two variables $\boldsymbol{\xi}$ and $\boldsymbol{T}$ in space and time have totally different physical units, individual step sizes are needed for efficient parameter updates.}
For the gradient descent, the key is to obtain the partial derivatives $\partial J^* / \partial \boldsymbol{\xi}$ and $\partial J^* / \partial \boldsymbol{T}$, which stand for the sensitivity of the optimal cost of \eqref{prob: our lower level formulation} to $\boldsymbol{\xi}$ and $\boldsymbol{T}$, respectively. We use OptNet~\cite{Amos20171} to obtain these two gradients. Specifically, the OptNet allows computing the analytical gradient of the optimal cost with respect to the coefficients of a QP (e.g., $P(\boldsymbol{T})$, $\boldsymbol{q}(\boldsymbol{T})$, $C(\boldsymbol{T})$, $A(\boldsymbol{T})$, $\boldsymbol{\xi}$, and $\boldsymbol{b}$ in \eqref{prob: our lower level formulation}) by implicitly differentiating the KKT condition. Using the method enabled by OptNet, we can directly obtain
\begin{equation}
    \frac{\partial J^*}{\partial \boldsymbol{\xi}}, \frac{\partial J^*}{\partial P(\boldsymbol{T})}, \frac{\partial J^*}{\partial \boldsymbol{q}(\boldsymbol{T})}, \frac{\partial J^*}{\partial C(\boldsymbol{T})}, \frac{\partial J^*}{\partial A(\boldsymbol{T})}.
\end{equation}
Using chain rule, we can obtain
\begin{align}
    \frac{\partial J^*}{\partial \boldsymbol{T}} = 
    & \frac{\partial J^*}{\partial P(\boldsymbol{T})} \frac{\partial P(\boldsymbol{T})}{\partial \boldsymbol{T}}
    + \frac{\partial J^*}{\partial \boldsymbol{q}(\boldsymbol{T})}\frac{\partial \boldsymbol{q}(\boldsymbol{T})}{\partial \boldsymbol{T}} \nonumber \\
    & +  \frac{\partial J^*}{\partial C(\boldsymbol{T})}  \frac{\partial C(\boldsymbol{T})}{\partial \boldsymbol{T}}
    + \frac{\partial J^*}{\partial A(\boldsymbol{T})}  \frac{\partial A(\boldsymbol{T})}{\partial \boldsymbol{T}}, \label{eq: partial derivative of cost to time assignment}
\end{align}
where $\frac{\partial P(\boldsymbol{T})}{\partial \boldsymbol{T}}$, $\frac{\partial \boldsymbol{q}(\boldsymbol{T})}{\partial \boldsymbol{T}}$, $\frac{\partial C(\boldsymbol{T})}{\partial \boldsymbol{T}}$, $\frac{\partial A(\boldsymbol{T})}{\partial \boldsymbol{T}}$ are easy to compute since $P(\boldsymbol{T})$, $\boldsymbol{q}(\boldsymbol{T})$, $C(\boldsymbol{T})$, $A(\boldsymbol{T})$ are explicit functions of~$\boldsymbol{T}$. We summarize our solution method to \eqref{prob: our formulation} in Alg.~\ref{alg: our algorithm}.

\setlength{\textfloatsep}{1pt}
\begin{algorithm}[t]
      \caption{Solution method of \eqref{prob: our formulation}} 
      \label{alg: our algorithm}
      \begin{algorithmic}[1]
            \Require Initial temporal assignment $\boldsymbol{T}_0$, initial spatial assignment $\boldsymbol{\xi}_0$, spatial step size $\alpha_1$, temporal step size $\alpha_2$, and termination condition $\mathcal{C}$.
            \Ensure Optimal spatial and temporal assignment  $(\boldsymbol{\xi}^*,\boldsymbol{T}^*)$ and polynomial coefficients $\boldsymbol{\sigma}^*$.
            \State $(\boldsymbol{\xi},\boldsymbol{T})  \gets (\boldsymbol{\xi}_0,\boldsymbol{T}_0 )$ 
            \While{$\mathcal{C}$ is FALSE}
                    \State Solve the~\eqref{prob: our lower level formulation} using a QP solver (with the parameters $P(\boldsymbol{T})$, $\boldsymbol{q}(\boldsymbol{T})$, $A(\boldsymbol{T})$, $\boldsymbol{b}$, $C(\boldsymbol{T})$, $\boldsymbol{\xi}$) to obtain the optimal solution~$\boldsymbol{\sigma}^*$ and optimal value~$J^*$
                    \State Obtain the gradients $\partial J^* / \partial \boldsymbol{\xi}$ and $\partial J^* / \partial \boldsymbol{T}$ using OptNet~\cite{Amos20171} and chain rule in~\eqref{eq: partial derivative of cost to time assignment}
                    \State Use the projected gradient descent in~\eqref{eq: projected gradient descent to update the parameters} to update $\boldsymbol{\xi}$ and $\boldsymbol{T}$ . 
                    \State $(\boldsymbol{\xi}^*,\boldsymbol{T}^*)  \gets (\boldsymbol{\xi},\boldsymbol{T})$
               \EndWhile
               \State \Return $(\boldsymbol{\xi}^*,\boldsymbol{T}^*)$ and $\boldsymbol{\sigma}^*$. 
    \end{algorithmic}
 \end{algorithm}

\section{Simulation results}\label{sec: simulation}
All the simulations shown in this section are implemented using MATLAB R2021b \revision{on a computer with a 2.20 GHz Intel i7-8750H CPU.} \revision{We use \texttt{quadprog} as the QP solver, in which we choose the interior-point method instead of the active-set method in \texttt{quadprog} because the former provided better solution quality with shorter computation time than the latter in our simulations. This observation is consistent with the solver comparison shown in~\cite{Sun20201}.} We compare our method with that of \cite{Sun20201} (referred to as ``compared method'' in this section) in three experiments. In the first one, we fix the number of adjustable waypoints and show a breakdown of the computation time. In the second one, we conduct a scalability experiment to show how the computation time scales with the number of adjustable waypoints. 
In the third one, we fix the number of adjustable waypoints and include the dynamic constraints.
Note that the number of adjustable waypoints equals the number of safety sets as indicated in~\eqref{eq: introduce adjustable waypoint}.

\subsection{Trajectory Planning with Two Adjustable Waypoints}

\begin{figure}[h]
\centering
\includegraphics[width = \columnwidth]{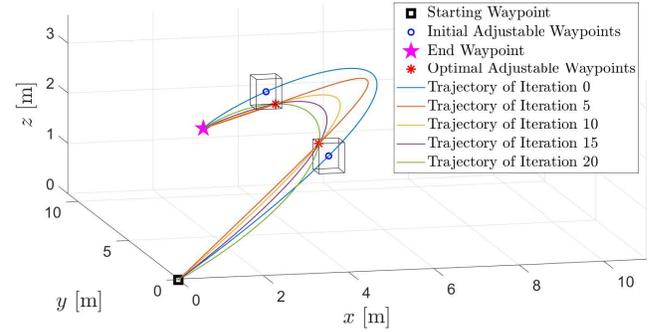}
\caption{An example with four random waypoints, where two middle ones are adjustable.}
\label{fig:traj_1}
\end{figure}

We randomly select four waypoints, forming a path with a total length of 10 m. Initially, we allocate 2 s to each trajectory segment. The start and end points are fixed with velocity, acceleration, jerk, and snap set to  0. 
Two adjustable waypoints are constrained in safety sets $\mathcal{C}_1$ and $\mathcal{C}_2$, which are cubes with 0.6 m side lengths centered at the initial waypoints. The scenario is illustrated in Fig.~\ref{fig:traj_1}.
Our method optimizes the time allocation and spatial assignment for the two adjustable waypoints, whereas the compared method only updates the time allocation (the adjustable waypoints are solved in their lower-level problem). 
The iterations terminate when the relative cost reduction between consecutive iterations is within 3\%. 

\begin{figure}[h]
\centering
\includegraphics[width = \columnwidth]{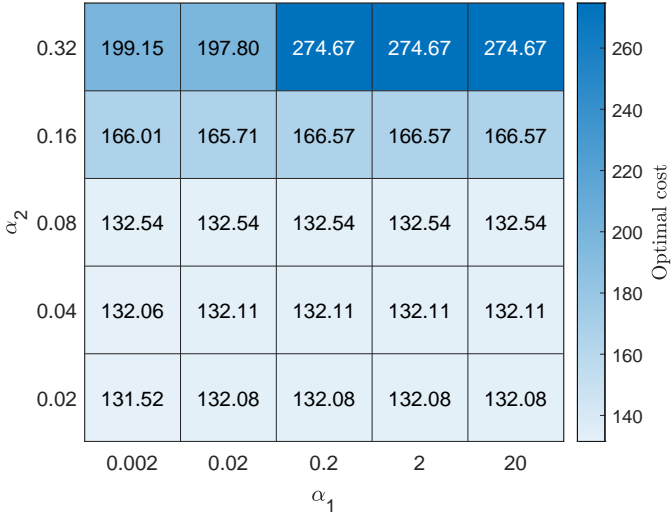}
\caption{Optimal costs of the trajectories planned by the proposed method under the combination of different step sizes $\alpha_1$ (spatial) and $\alpha_2$ (temporal)}
\label{fig:hm_fincost}
\end{figure}

\begin{figure}[h]
\centering
\includegraphics[width = \columnwidth]{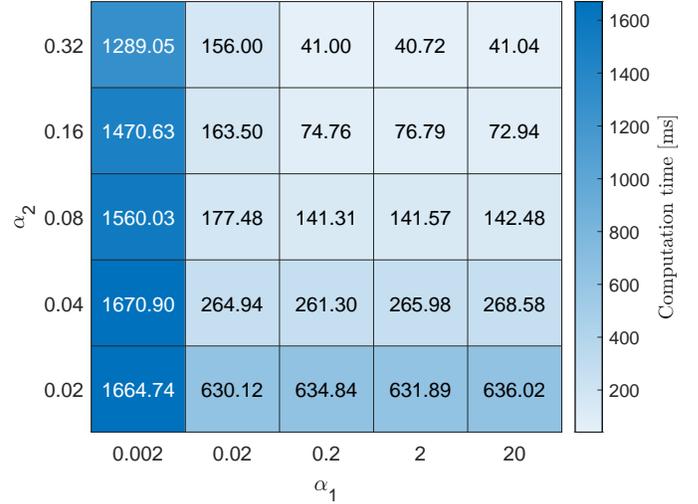}
\caption{Total computation time of the trajectory planned by the proposed method under the combination of different step sizes $\alpha_1$ (spatial) and $\alpha_2$ (temporal)}
\label{fig:hm_comptime}
\end{figure}

\revision{The hyperparameter choice of the spatial and temporal step sizes, $\alpha_1$ and $\alpha_2$, largely determine the performance and computation time of the proposed method. We first investigate the optimal cost and total computation time subject to different combinations of $\alpha_1$ and $\alpha_2$, where the results are shown in Figs.~\ref{fig:hm_fincost} and~\ref{fig:hm_comptime}. 
It can be observed that when $\alpha_1 \geq 0.2$, increasing $\alpha_1$ does not lower computation time, and when $\alpha_2 \leq 0.08$, decreasing $\alpha_2$ does not reduce the optimal cost significantly. Therefore, we choose $\alpha_1=0.2$ and $\alpha_2=0.08$ to balance the optimality and computation time, where we  use this combination of step sizes in the simulations next.}

\begin{figure}[h]
\centering
\includegraphics[width = \columnwidth]{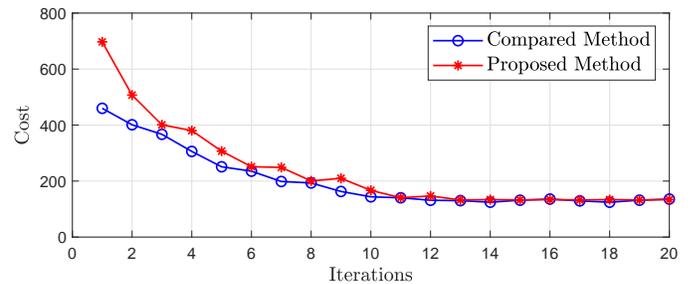}
\caption{Cost reduction through iterations in the case with two adjustable waypoints}
\label{fig:vali_cost}
\end{figure}

Figure~\ref{fig:traj_1} shows the gradually updated trajectory in the first 20 iterations. Compared to the stretched initial trajectory, the final trajectory in iteration 20 is better shaped, indicating the benefit of simultaneously conducting spatial and temporal assignments. Meanwhile, the adjustable waypoints are contained within the cubic safety sets and close to the initial waypoints. The cost reduction is demonstrated in Fig.~\ref{fig:vali_cost}, where both the proposed method and compared method are initialized with identical temporal assignments and terminate on the 13th iteration according to the 3\% relative cost reduction. It can be seen that both methods achieve similar optimal costs in a similar number of iterations.

The optimal cost and computation time of the two methods are shown in Table~\ref{table:cost and time}. The comparison shows that the proposed method reaches a slightly smaller cost with a significant time reduction compared to \cite{Sun20201} with the same iterations. We dissect the computation in each iteration into four parts and count the individual computation time. The averaged computation time \revision{and its standard deviation} is shown in Table~\ref{table:time details}. \revision{In the compared method, the computation time of solving QP fluctuates due to the inequality constraints in the QPs. Specifically, the change in the time coefficients dramatically affects the parameters of the QP problem, causing the number of iterations and computation time to vary in a wide range for solving the QP.} Since the compared method updates the adjustable waypoints in the lower-level QP, the spatial update time is not applicable. It can be observed that the large gap in total computation time mainly comes from the discrepancy in the time of solving the QP in these two methods. Besides, we observe that the QP solver's reported iterations are consistently 1 when running the proposed method, whereas the number could vary from 5 to 20 when running the compared method. These observations are consistent with our previous analysis that the merits of the proposed method mainly lie in the exclusion of inequality constraints when solving QP. 


\begin{table}[h]
\small
\captionsetup{font=small}
\caption{Comparison on the optimal costs and computation time till convergence}
\label{table:cost and time}
\centering
\begin{tabular}{l c c}
\toprule
      & Compared & Proposed   \\ \midrule
Optimal Cost                & 135.69    & \textbf{132.54}\\ 
Total Computation Time {[}ms{]} & 361.14    & \textbf{141.31}\\ \bottomrule
\end{tabular}
\end{table}

\begin{table}[h]
\small
\captionsetup{font=small}
\caption{Breakdown of the averaged computation time and its standard deviation for one iteration. (U) and (L) associate with the upper- and lower-level problems, respectively. }
\label{table:time details}
\centering
\begin{tabular}{lrr}
\toprule

Computation Time [ms] & Compared      & Proposed \\ \midrule
 Gradients from OptNet (U)            & 5.21$\pm$0.45    & {\textbf{4.04$\pm$0.03}}                 \\
Update QP Parameters (U)       	 & 1.75$\pm$0.08     & {\textbf{1.72$\pm$0.12}}              \\
Update Waypoints (U)                 	  & {N/A}      & 0.34$\pm$0.03             \\ 
Solving QP (L)                            	   & 20.82$\pm$17.39      & {\textbf{4.77$\pm$0.29}}              \\
\midrule
Total            & 27.78$\pm$17.92   & {\textbf{10.87$\pm$0.47}}          \\ \bottomrule
\end{tabular}
\end{table}

\revision{We also study how the size of the safety sets affects the performance of the two methods. Keeping all other conditions fixed, we expand the side length of the two cubic safety sets to 0.9 m and 1.2 m.  With new safety sets, we conduct the simulations and record both methods' optimal cost and computation time, which are shown in Table~\ref{table:variable safety sets}. The results show that for both methods, the optimal cost and total computation time decline remarkably when the side length of the safety sets expands from 0.6 m to 0.9 ~m. However, the values remain almost unchanged when the side length of safety sets expands from 0.9 m to 1.2 m. It occurs because the optimal waypoints are on the boundary of the cubes with 0.6 m side length, in contrast to those in the interior of the cubes with 0.9 m side length. In other words, the optimization constraints change from active to inactive when the side length of safety sets sides expands from 0.6 m to 0.9 m. Therefore, expanding from 0.6 m to 0.9 m lowers the cost and computation time dramatically. However, expanding from 0.9 m to 1.2 m does not have the same effect since the optimal waypoints are all in the interior of the safety sets. Regardless of the size of safety sets, the proposed method has the advantage of shorter computation time while achieving almost the same planning quality as the compared method for all three cases. }

\begin{table}[h]

\vspace{0.2cm}
\small
\captionsetup{font=small}
\caption{Comparison on optimal cost and computation time for safety sets with different sizes}
\label{table:variable safety sets}
\centering

\begin{tabular}{clrr}
\toprule
\begin{tabular}[c]{@{}c@{}}Side Length of \\ Safety Sets [m]\end{tabular} &                                 & Compared & Proposed \\
\midrule
\multirow{2}{*}{0.6}                                                    & Optimal Cost                    & 135.69          & \textbf{132.54}          \\
                                                                  &  Comp. Time {[}ms{]} & 361.14       & \textbf{141.31}\\
\midrule
\multirow{2}{*}{0.9}                                                    & Optimal Cost                    & \textbf{98.53}           & 99.16           \\
                                                                        &  Comp. Time {[}ms{]} & 261.30          & \textbf{132.30} \\
\midrule
\multirow{2}{*}{1.2}                                                    & Optimal Cost                    & 96.44           & \textbf{95.25}           \\
                                                                        &  Comp. Time {[}ms{]} & 283.62          & \textbf{159.48}\\         
\bottomrule
\end{tabular}

\vspace{-0.5cm}
\end{table}

\subsection{Scalability Experiment with Multiple Adjustable Waypoints  }

\begin{figure}[h]
\centering
\includegraphics[width = \columnwidth]{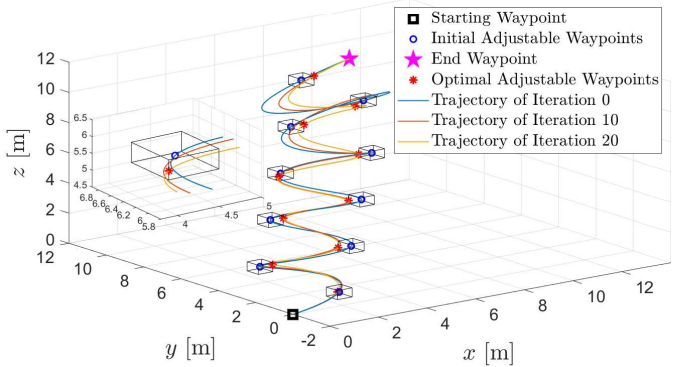}
\caption{An example with twelve random waypoints, where ten middle ones are adjustable.}
\label{fig:traj_2}
\end{figure}

We conduct the scalability experiment to test the scalability of the proposed method and also demonstrate the proposed method’s merit in computation time when planning more complex trajectories. In this experiment, we conduct a series of tests with the number of adjustable waypoints spanning from one to ten. The start and end waypoints are fixed at (0, 0, 0) and (10, 10, 10), respectively, with velocity, acceleration, jerk, and snap set to 0 at both waypoints. The adjustable waypoints are initialized as the vertices of a slalom path and are evenly distributed in the $z$-axis (see Fig.~\ref{fig:traj_2} for an illustration with ten adjustable waypoints). Therefore, the planned trajectories twist and turn around the line connecting the start and end waypoints, causing a dramatic change in the cost as the number of adjustable waypoints increases (the cost increases from 20 to 833,536 as the waypoints increase from one to ten). The total time of the trajectory is 8~s. The initial time allocated to the adjustable waypoints is equally distributed in the 8~s interval. For both methods, the adjustable waypoints are constrained within a cube with 0.6~m side length centered at the initially selected waypoints. For all cases, the termination criterion is set to 5\% relative reduction (which is met within 20 iterations for both methods). Figure~\ref{fig:traj_2} shows an example with ten adjustable waypoints. Similar to what has been observed in Fig.~\ref{fig:traj_1}, the final trajectory in iteration 20 is better shaped and less stretched than the initial trajectory.

\begin{figure}[h]
\centering
\includegraphics[width = \columnwidth]{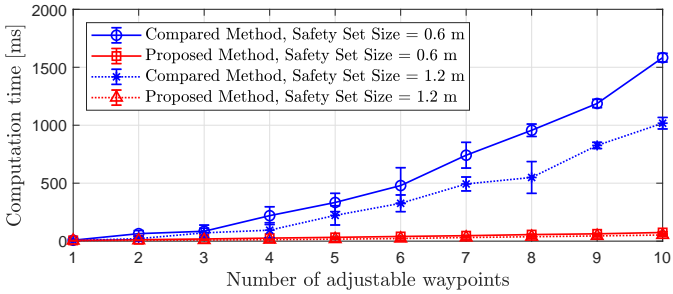}
\caption{Average computation time in one iteration with error bars showing the standard deviations.}
\label{fig:scal_comptime}

\end{figure}

The total computation time \revision{(when the  5\% relative cost reduction is reached)} of the two methods scales similar to that of a single iteration shown in Table~\ref{table:time details}. The time comparison is demonstrated in Fig.~\ref{fig:scal_comptime}. It is clear that the proposed method has a great advantage compared with \cite{Sun20201}: The computation time of the proposed method scales linearly with the number of adjustable waypoints, whereas the compared method in \cite{Sun20201} scales exponentially. \revision{Specifically, we conduct a regression analysis of the computation time of the two methods with the safety set size of 0.6 m. Denote the computation time by $t$. Then for $N \in \{1,2,\dots,10\}$ number of safety sets, the proposed method's computation time fits $t = 7.52N-3.42$ with an R-square score of 0.99 (the exponential regression $t = 11.45 \exp{(0.19N)}$ gives an R-score of 0.97, which has less fidelity than the linear fit). The compared method's computation time fits $t = 72.06 \exp{(0.31N)}$ with an R-square score of 0.98 (the linear  regression $t = 170.32N-370.94$ gives an R-score of 0.93, which has less fidelity than the exponential fit).} 
\revision{Furthermore, the standard deviations are shown in Fig.~\ref{fig:scal_comptime} by error bars, which indicate the more consistent computation time of the proposed method than the compared method.}

\begin{figure}[h]
\centering
\includegraphics[width = \columnwidth]{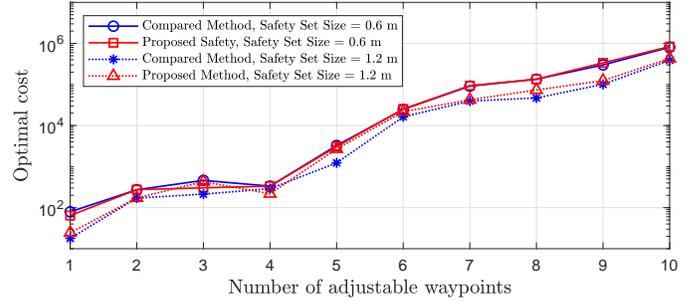}
\caption{Comparison of the optimal cost as the number of adjustable waypoints increases.}
\label{fig:scal_cost}

\end{figure}

The optimal costs of the two methods are shown in Fig.~\ref{fig:scal_cost}. The proposed method either reaches a smaller cost or is within 20\% of the cost obtained by the compared method. 
Summarizing the comparison in Fig.~\ref{fig:scal_comptime} and Fig.~\ref{fig:scal_cost}, we conclude that the proposed method can deliver similar performance in the planned trajectory with the benefit of significantly reducing the computation time.  

\revision{We further compare the two methods subject to larger safety sets. By expanding the side length of safety sets from 0.6 m to 1.2 m and keeping all other conditions unchanged as above, we display the average computation time in one iteration and optimal cost of both methods in Fig.~\ref{fig:scal_comptime} and Fig.~\ref{fig:scal_cost}. The results show that the computation time and the optimal cost of both methods decrease with the expansion of safety sets. Meanwhile, under larger safety sets, the proposed method still has the advantage of generating trajectories with similar quality while significantly reducing the computation time than the compared method.}

\revision{
\subsection{Trajectory Optimization with Dynamic Constraints}}

\begin{figure}[h]
\centering
\includegraphics[width = \columnwidth]{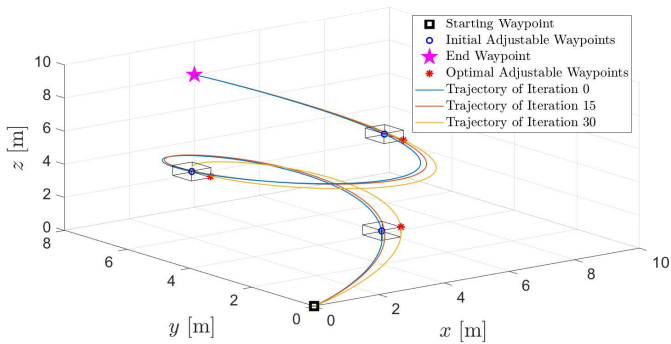}
\caption{An example with five random waypoints subject to dynamic constraints, where the three middle ones are adjustable }
\label{fig:traj_3}

\end{figure}

\revision{We randomly select five waypoints, forming a path with a total length of 20 m. Initially, we allocate 2 s to each trajectory segment. The start and end points are fixed with velocity, acceleration, jerk, and snap set to 0. Three adjustable waypoints are constrained in the cubic safety sets with 0.6~m side lengths centered at the initial waypoints. Meanwhile, the maximum velocity and acceleration are set to 5 m/s and 5~m/s$^2$, respectively, on the three adjustable waypoints. The scenario is illustrated in Fig.~\ref{fig:traj_3}.}

\begin{figure}[h]
\centering
\includegraphics[width = \columnwidth]{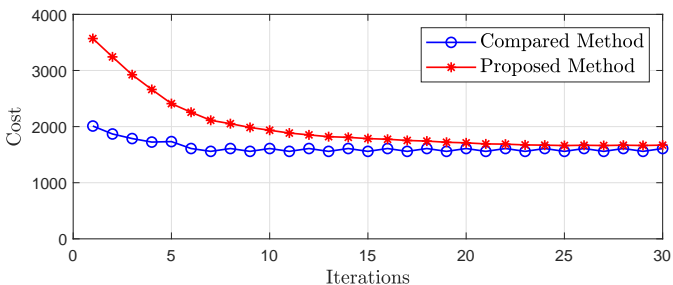}
\caption{Cost reduction through iterations with three adjustable waypoints subject to dynamic constraints}
\label{fig:dyn_cost}

\end{figure}

\revision{Our method optimizes the time allocation and the augmented spatial assignment for the three adjustable waypoints in the upper-level problem, whereas the compared method updates the time allocation only in their upper-level problem and solves the adjustable waypoints velocity and acceleration subject to the maximum in their lower-level problem.}
\revision{The cost reduction is demonstrated in Fig.~\ref{fig:dyn_cost}, where both the proposed method and compared method are initialized with identical temporal assignments. Under the termination condition of 3\% relative cost reduction, the compared method terminates at the 6th iteration, and the proposed method terminates at the 25th iteration. Meanwhile, the proposed method's optimal cost is within 3.5\% of the cost obtained by the compared method.}

\revision{Table~\ref{table:dyncon cost and time} shows the optimization results of two methods. Though the proposed method takes more iterations to terminate, the remarkable advantage in computation time of one iteration still makes the proposed method use much less time to plan a trajectory in contrast with the compared method, while a similar optimality is achieved in both methods. The breakdown of the average computation time for one iteration and their standard deviations are demonstrated in Table~\ref{table:dyncon time details}. It is clear that the major gap between the two methods in computation time is caused by the gap in solving the lower-level QP. For each adjustable waypoint, the compared method needs to solve 18 inequality constraints while the proposed method delegates them to the upper-level problem and solves the lower-level problem with equality constraints only, reducing the time consumption significantly.}

\begin{table}[h]
\small
\captionsetup{font=small}
\caption{Comparison on the optimal costs and computation time till convergence when the dynamic constraints are present.}
\label{table:dyncon cost and time}
\centering
\begin{tabular}{l c c}
\toprule
      & Compared & Proposed   \\ \midrule
Optimal Cost                & \textbf{1609.34}    & 1662.36\\ 
Total Computation Time {[}ms{]} & 4342.73    & \textbf{362.75}\\ \bottomrule
\end{tabular}
\end{table}

\begin{table}[H]
\small
\captionsetup{font=small}
\caption{Breakdown of the averaged computation time and its standard deviation for one iteration when the dynamic constraints are present. (U) and (L) associate with the upper- and lower-level problems, respectively. }
\label{table:dyncon time details}
\centering
\begin{tabular}{lrr}
\toprule

Computation Time [ms] & Compared      & Proposed \\ \midrule
 Gradients from OptNet (U)            & 9.82$\pm$1.49    & {\textbf{5.56$\pm$0.88}}                 \\
Update QP Parameters (U)       	 & {\textbf{1.73$\pm$0.21}}     & 2.41$\pm$0.61         \\
Update Waypoints (U)                 	  & {N/A}      & 0.35$\pm$0.03             \\ 
Update Dynamic Constraints(U) &{N/A} & 0.38$\pm$0.04 \\
Solving QP (L)                            	   & 608.84$\pm$19.54      & {\textbf{5.81$\pm$0.29}}              \\
\midrule
Total            & 620.39$\pm$21.24   & {\textbf{14.51$\pm$1.85}}          \\ \bottomrule
\end{tabular}
\end{table}

\section{Conclusion} \label{sec: conclusion}
We present a novel method of UAV trajectory planning based on bilevel optimization by simultaneously conducting spatial and temporal assignments. Our bilevel optimization is composed of a lower-level problem that is a QP with equality constraints only (which contributes majorly to the reduced computation time) and an upper-level problem that uses analytical gradients to make the spatial and temporal update on the adjustable waypoints. Simulation results show that our method has great advantages in computation time compared to the existing bilevel optimization method, where the former scales linearly and the latter scales exponentially to the number of adjustable waypoints. Future work will deploy our method on a real quadrotor to demonstrate its capability for onboard and real-time trajectory planning in complex environments.

\bibliographystyle{IEEEtran}
\bibliography{ref}
\end{document}